# Democratizing planetary-scale analysis: An ultra-lightweight Earth embedding database for accurate and flexible global land monitoring

Shuang Chen[1], Jie Wang[2], Shuai Yuan[1], Jiayang Li[2], Yu Xia[2], Yuanhong Liao[3], Junbo Wei[2], Jincheng Yuan[2], Xiaoqing Xu[2], Xiaolin Zhu[4], Peng Zhu[1], Hongsheng Zhang[1], Yuyu Zhou[1], Haohuan Fu[5], Huabing Huang[6], Bin Chen[7,8], Fan Dai[8], Peng Gong[1,8,9]*

[1] Department of Geography, The University of Hong Kong, Hong Kong, China
[2] Pengcheng Laboratory, Shenzhen 518000, China
[3] Ministry of Education Key Laboratory for Earth System Modeling, Department of Earth System Science, Tsinghua University, Beijing 100084, China
[4] Department of Land Surveying and Geo-Informatics, The Hong Kong Polytechnic University, Hong Kong, China
[5] Tsinghua Shenzhen International Graduate School, Shenzhen, China
[6] School of Geospatial Engineering and Science, Sun Yat-sen University, Guangzhou 510275, China
[7] Department of Architecture, The University of Hong Kong, Hong Kong, China.
[8] Institute for Climate and Carbon Neutrality, The University of Hong Kong, Hong Kong, China
[9] Department of Earth Sciences, The University of Hong Kong, Hong Kong, China

*Correspondence to: Jie Wang (wangj10@pcl.ac.cn) and Peng Gong (penggong@hku.hk)

**Abstract.** The rapid evolution of satellite-borne Earth Observation (EO) systems has fundamentally revolutionized terrestrial monitoring, yielding comprehensive petabyte-scale archives. However, the immense computational resources and storage volumes required for global-scale analysis often preclude widespread use by many research teams, hindering broader scientific adoption and the execution of planetary-scale studies. To address these barriers, we present the Embedded Seamless Data (ESD), an ultra-lightweight, 30-m global Earth embedding database spanning the 25-year period from 2000 to 2024. By transforming high-dimensional, multi-sensor observations from the Landsat series (5, 7, 8, and 9) and MODIS Terra into information-dense, quantized latent vectors, ESD distills essential geophysical and semantic features into a unified latent space. Utilizing the ESDNet architecture and Finite Scalar Quantization (FSQ), the dataset achieves a transformative ~340-fold reduction in data volume compared to raw daily archives. This compression allows the entire global land surface for a single year to be encapsulated within approximately 2.4 TB, enabling decadal-scale global analysis on standard local workstations. Rigorous validation demonstrates that ESD maintains high reconstructive fidelity across the spectral dimension, achieving a mean Absolute Error (MAE) of 0.0130, a Root Mean Square Error (RMSE) of 0.0179, and a Correlation Coefficient (CC) of 0.8543. By condensing the annual phenological cycle into 12 temporal latent steps, the embeddings provide inherent denoising effects and a semantically organized latent space that outperforms raw reflectance data in downstream land-cover classification tasks, achieving an overall accuracy of 79.74% compared to 76.92% using raw sensor fusion data. With robust few-shot learning capabilities and longitudinal consistency across 25 years, the ESD product provides a versatile foundation for democratizing planetary-scale Earth system research and advancing next-generation geospatial artificial intelligence.

# 1. Introduction

Over the past half-century, satellite-borne Earth Observation (EO) systems have fundamentally revolutionized our capacity to monitor terrestrial processes at a global scale (Drusch et al., 2012; Markham and Helder, 2012; Morisette et al., 2002; Wulder et al., 2022), yielding comprehensive datasets with unprecedented spatial and temporal continuity. These remotely sensed data have become pivotal to Earth system science, facilitating quantitative assessments of land surface dynamics (Brown et al., 2022; Gong et al., 2019, 2013; Liu et al., 2021; Zanaga et al., 2022), phenological cycles (Bolton et al., 2020; Piao et al., 2019), forest cover change (Aguirre-Gutiérrez et al., 2022; Forzieri et al., 2022; Hansen et al., 2013), inland water dynamics (Ji et al., 2018; Pekel et al., 2016; Pickens et al., 2022, 2020), and patterns of urban expansion (Gong et al., 2020, 2012; Huang et al., 2022).

Recent decades have seen the increasing availability of high-spatiotemporal-resolution satellite imagery and high-performance computing platforms, ushering in a new era for global-scale remote sensing applications (Drusch et al., 2012; Gorelick et al., 2017; Morisette et al., 2002; Wulder et al., 2022). For instance, coarse-spatial-resolution sensors, such as AVHRR and MODIS, have provides long-term records essential for climate variability and vegetation monitoring (Boyte et al., 2018; Chuvieco et al., 2005; Huete et al., 2002; Xu et al., 2022). Simultaneously, the medium-resolution Landsat missions have delivered the most extensive continuous record of land surface reflectance to date (Wulder et al., 2022). The adoption of a free and open data policy in 2008 (Woodcock et al., 2008) further accelerated the study of land-use change and ecosystem dynamics. More recently, the Sentinel-2 constellation has enhanced revisit frequencies, spatial resolutions, and spectral sampling, revealing new frontiers in precision agriculture and fine-scale land-cover mapping (Drusch et al., 2012).

Despite these advancements, many land monitoring applications necessitate EO data that possess both high spatial and high temporal resolutions—a combination that remains unattainable from any single satellite sensor (Zhu et al., 2018). To bridge this gap, various multi-sensor fusion algorithms and datasets have been developed (Chen et al., 2024, 2023; Claverie et al., 2018; Gao et al., 2006; Goyena et al., 2023; Guo et al., 2020; Liu et al., 2022; Shang and Zhu, 2019; Zhu et al., 2016, 2010). These methods have enabled innovative applications in monitoring forest disturbances (Boyte et al., 2018; Hansen et al., 2008; Hilker et al., 2009; Shang et al., 2022) and surface water dynamics (Abowarda et al., 2021; Chen et al., 2018; Declaro and Kanae, 2024). Nevertheless, this rapid proliferation of sensor diversity and spatial-temporal-spectral resolution has resulted in an explosion of data volume, posing significant challenges for the storage, management, and processing of large-scale applications (Tuia et al., 2025).

While cloud-based platforms like Google Earth Engine and Microsoft Planetary Computer have democratized access to petabyte-scale datasets and scalable workflows (Gorelick et al., 2017; Smits, 2022; Zhao et al., 2021), significant barriers remain. High computational costs, processing latencies, and associated financial constraints often preclude widespread use by smaller research teams, thereby hindering broader adoption and the execution of global-scale analyses (Tuia et al., 2025).

In parallel, artificial intelligence—specifically deep learning—has catalysed a paradigm shift in remote sensing data analysis (Tuia et al., 2025), achieving state-of-the-art performance in tasks such as land-cover classification (Ienco et al., 2019), semantic segmentation (Wieland et al., 2023), and change detection (Soto Vega et al., 2021). The recent advent of self-supervised pretraining (Devlin et al., 2019; He et al., 2022) and geospatial-specific foundation models has demonstrated remarkable generalization across multi-sensor and multi-modal EO inputs (Brown et al., 2025; Cong et al., 2022; Guo et al., 2024; Jakubik et al., 2023; Manas et al., 2021; Szwarcman et al., 2025). Nevertheless, the architectural complexity of these remote sensing foundation models—which frequently encompass billions of parameters—presents significant challenges. When the massive scale of these models is compounded by the vast data volumes inherent to global-scale remote sensing, the resulting computational costs for both pretraining and inference can become prohibitive. Consequently, despite their high performance and promise, these models remain difficult to deploy in many practical and operational settings due to the substantial resource requirements and the inherently immense size of remote sensing archives (Tuia et al., 2025).

Earth embeddings offer a promising solution by transforming high-dimensional, multi-modal satellite observations into compact, information-dense numerical vectors (Brown et al., 2025; Czerkawski et al., 2024; Feng et al., 2025; Klemmer et al., 2025). By projecting raw spectral data into a unified latent space, these embeddings distill essential geophysical and semantic features while significantly reducing data dimensionality (Brown et al., 2025). This approach not only mitigates the storage and processing bottlenecks inherent in petabyte-scale archives but also facilitates efficient downstream analysis—including cross-sensor fusion, classification, and change detection—without requiring the resource-intensive re-deployment of massive foundation models for every task (Brown et al., 2025; Klemmer et al., 2025).

Despite these theoretical advantages, there remains a notable scarcity of global-scale embedding datasets ready for immediate use by the broader scientific community (Klemmer et al., 2025). While pioneering initiatives such as Major TOM (Czerkawski et al., 2024), AlphaEarth Foundation (Brown et al., 2025), and TESSERA (Feng et al., 2025) represent a foundational precedent for the dissemination of large-scale geospatial embeddings, existing databases face critical limitations for operational Earth system monitoring (Klemmer et al., 2025). Specifically, next-generation embedding databases should provide: **(i) enhanced compression ratios** to further mitigate the storage and computational constraints of global-scale applications (Klemmer et al., 2025; Tuia et al., 2025); **(ii) extended temporal coverage** to support decadal-scale longitudinal studies; and **(iii) the explicit preservation of temporal structures** to characterize vital intra-annual dynamics and phenological variations (Klemmer et al., 2025)**.**

To bridge these gaps, we present the Embedded Seamless Data (ESD), a global-scale 30-m Earth embedding dataset designed for accurate and efficient environmental monitoring. Derived from a fusion of Landsat-5/7/8/9 and MODIS Terra observations spanning from 2000 to 2024, ESD offers several transformative advantages over existing products. First, it achieves an exceptionally high compression ratio; the entire global land surface for a single year is encapsulated within approximately 2.4 TB, representing a significant reduction in storage requirements that democratizes petabyte-scale analysis for researchers with limited computational infrastructure. Second, ESD explicitly preserves temporal structures and information, providing the latent depth necessary to resolve intra-annual dynamics and complex phenological cycles. Third, the high fidelity of these embeddings allows for the reconstruction of near-daily 30-m observations with remarkably low error rates, providing a continuous proxy for surface processes. Finally, ESD provides extended temporal coverage spanning from 2000 to 2024, offering the longitudinal consistency required to track decadal shifts in Earth system processes.

The remainder of this paper is organized as follows: **Section 2** describes the multi-source satellite observations used to develop and validate the ESD dataset. **Section 3** outlines the methodological framework for the construction of the embeddings and the protocols used for accuracy assessment. In **Section 4**, we present the comprehensive validation results alongside several case studies that demonstrate the dataset's utility in monitoring land surface dynamics. **Section 5** discusses the technical advantages of ESD, its inherent limitations, and its potential for future Earth system research. Finally, **Section 6** provides details on data availability and access through open-science repositories, followed by concluding remarks in **Section 7**.

## 2. Materials

### 2.1 Input satellite imagery and auxiliary dataset

2.1.1 Global 30m Seamless Data Cube of Land Surface Reflectance

The development of the Embedded Seamless Data (ESD) is founded upon the integration of two primary satellite constellations: the Landsat series (5, 7, 8, and 9) and MODIS (Terra). As listed in **Table 1**, we utilize six spectral bands—blue, green, red, near-infrared (NIR), and two shortwave infrared (SWIR1 and SWIR2)—which provide the necessary spectral range for comprehensive environmental monitoring and land-surface characterization.

To generate a gap-free foundation for our embeddings, we first synthesized these multi-sensor observations into the Global 30 m Seamless Data Cube (SDC30) of Land Surface Reflectance (Chen et al., 2024). This was achieved through a rigorous processing pipeline encompassing data harmonization, missing-value reconstruction, and spatiotemporal fusion. The resulting SDC30 provides a spatially consistent, 30 m resolution record of global surface reflectance spanning the 25-year period from 2000 to 2024. This seamless data cube serves as the direct input for the construction of the ESD, transforming the reconstructed reflectance values into high-efficiency, information-dense latent representations.

**Table 1: Attributes of the six employed spectral bands from Landsat 5 TM, 7 ETM+, 8-9 OLI, and MODIS Terra products (Markham and Helder, 2012; Masek et al., 2020; Morisette et al., 2002).**

|  | Wavelengths (micrometers) | | | | |
| --- | --- | --- | --- | --- | --- |
|  | Landsat 5 TM | Landsat 7 ETM+ | Landsat 8 OLI | Landsat 9 OLI-2 | MODIS Terra |
| Blue | 0.450-0.520 | 0.450-0.515 | 0.452-0.512 | 0.452-0.512 | 0.459-0.479 |
| Green | 0.520-0.600 | 0.525-0.600 | 0.533-0.590 | 0.532-0.589 | 0.545-0.565 |
| Red | 0.630-0.690 | 0.630-0.690 | 0.636-0.673 | 0.636-0.672 | 0.620-0.670 |
| NIR | 0.760-0.900 | 0.760-0.900 | 0.851-0.879 | 0.850-0.879 | 0.841-0.876 |
| SWIR1 | 1.550-1.750 | 1.550-1.750 | 1.566-1.651 | 1.565-1.651 | 1.628-1.652 |
| SWIR2 | 2.080-2.350 | 2.080-2.350 | 2.107-2.294 | 2.105-2.294 | 2.105-2.155 |
| **Spatial resolution** | 30-m | 30-m | 30-m | 30-m | 500-m (near) |
| **Revisit frequency** | 16-day | 16-day | 16-day | 16-day | Daily (near) |

2.1.2 NASADEM: NASA 30m Digital Elevation Model

To incorporate topographic context into the embeddings, we utilized the NASADEM Global Digital Elevation Model (GDEM) at a 30 m spatial resolution (NASA JPL, 2020). NASADEM represents a major refinement of the Shuttle Radar Topography Mission (SRTM) archive, achieving superior vertical accuracy and data continuity by synthesizing SRTM data with auxiliary high-quality elevation sources. Specifically, it integrates observations from the ASTER GDEM, the ICESat Geoscience Laser Altimeter System (GLAS), and the ALOS Panchromatic Remote-sensing Instrument for Stereo Mapping (PRISM) to eliminate voids and improve the representation of complex terrain. Within the ESD framework, this elevation layer provides essential terrain-dependent features—such as slope, aspect, and elevation—that act as critical static covariates. These features enhance the model's capacity to characterize land-surface processes and improve the performance of downstream applications in hydrological modeling, ecosystem mapping, and terrain-aware change detection.

**2.2 Supervisory land cover datasets**

To ensure that the learned latent representations within the ESD align with meaningful biophysical properties and high-level semantic categories, we incorporated several authoritative land cover products as supervisory signals during the model training phase. By constraining the embedding space with these thematic labels, the model is better equipped to distill essential geophysical features from raw reflectance, thereby enhancing the interpretability and performance of the embeddings in downstream classification and change-detection tasks.

2.2.1 Annual maps of global artificial impervious area

The Global Artificial Impervious Area (GAIA) dataset (Gong et al., 2020) was utilized to provide long-term supervision for built-up environments. Derived from the Landsat archive at a 30 m spatial resolution, GAIA provides annual global maps of impervious surfaces from 1985 to 2024. This dataset is characterized by high thematic accuracy (exceeding 90%) and is particularly effective in mitigating the spectral confusion between building shadows and open water bodies—a common challenge in urban remote sensing. In this study, GAIA serves as a temporal anchor to ensure the stability of the embeddings in rapidly urbanizing regions over the 25-year study period.

### 2.2.2 ESA WorldCover 2021

The ESA WorldCover 2021 dataset (Zanaga et al., 2022), a global land cover product at 10-meter spatial resolution, was employed as supervisory data in the training phase of this study. Generated from Sentinel-1 and Sentinel-2 satellite imagery, this dataset provides detailed and reliable land cover classifications encompassing multiple land cover categories, including forests, urban areas, croplands, water bodies, and various natural vegetation types. The high-resolution and globally consistent nature of ESA WorldCover 2021 made it an ideal supervisory dataset, significantly enhancing the semantic quality and interpretability of the generated embedding vectors.

### 2.2.3 GLAD global crop extent dynamics

The GLAD Global Crop Extent Dynamics (GLAD-CE) dataset (Potapov et al., 2021) was employed to supervise the identification of agricultural landscapes. This 30 m product provides a consistent time series of cropland extent from 2000 to 2019, partitioned into four-year intervals. Following the GLAD definition, cropland includes land used for annual and perennial herbaceous crops while excluding woody perennials and permanent pastures. The inclusion of GLAD-CE ensures that the resulting ESD embeddings accurately preserve the phenological signatures unique to intensive agriculture and crop rotation cycles (https://glad.umd.edu/dataset/croplands; last accessed: December 2, 2024).

### 2.2.4 GLAD global monthly surface water dynamics

To characterize intra-annual hydrological variations, we leveraged the Global Land Analysis and Discovery Global Surface Water Dynamics (GLAD-SW) dataset (Pickens et al., 2022, 2020). This product provides monthly global surface water extent at a 30 m resolution from 1999 to 2021. With reported user's and producer's accuracies of 93.7% and 96.0%, respectively, GLAD-SW provides the necessary temporal supervision to resolve seasonal and inter-annual fluctuations in aquatic ecosystems. The dataset was accessed via the official GLAD repository (https://glad.umd.edu/dataset/global-surface-water-dynamics; last accessed: December 2, 2024).

## 2.3 Validation datasets

### 2.3.1 FAST: First All-season Sample seTs for global land cover classification

The First All-Season Sample Set (FAST) serves as the primary reference for benchmarking our classification performance. FAST is an evolution of the Finer Resolution Observation and Monitoring of Global Land Cover (FROM-GLC) project (Gong et al., 2013), comprising 91,619 training and 36,636 validation locations worldwide.

The spatial allocation of validation samples followed a rigorous hex-grid approach, where the global land surface was partitioned into approximately 7,000 equal-area hexagons, with five pixel locations randomly selected within each. To capture phenological variability, the original single-date samples were expanded to a seasonal framework using primarily Landsat 8 OLI surface reflectance from 2014–2015 (Li et al., 2017). This seasonal expansion was manually verified by a team of expert interpreters using a multi-source evidence approach, integrating MODIS EVI time series, monthly climatology (temperature and precipitation), and high-resolution imagery from Google Earth.

The FAST classification system encompasses 11 primary categories: cropland, forest, grassland, shrubland, wetland, water, tundra, impervious surface, bareland, snow/ice, and cloud. For the purposes of this study, we aggregated seasonal labels into annual representations and excluded cloud-contaminated samples. To ensure spatial precision in the 30 m embedding process, only the central pixel of each sample unit was utilized for training and validation.

## 3. Methodology

### 3.1 Overview of the ESD production pipeline

This section outlines the integrated framework developed to generate the Embedded Seamless Data (ESD), transitioning from raw multi-sensor observations to information-dense latent representations. As illustrated in **Figure 1**, the pipelined ESDNet is organized into three primary modules:

(a) **Spatiotemporal Fusion**: The harmonization of heterogeneous Landsat and MODIS archives into the unified 30 m Seamless Data Cube (SDC30) (Chen et al., 2024).

(b) **Latent Encoding and Quantization**: A deep learning-based encoding process that transforms high-dimensional spectral-temporal features into quantized embedding vectors.

(c) **Multitask Inference and Reconstruction**: A suite of specialized task heads $\{D_1, D_2, D_3, D_4, ...\}$ that leverage the embeddings to generate downstream information products and reconstruct the original SDC30 inputs.

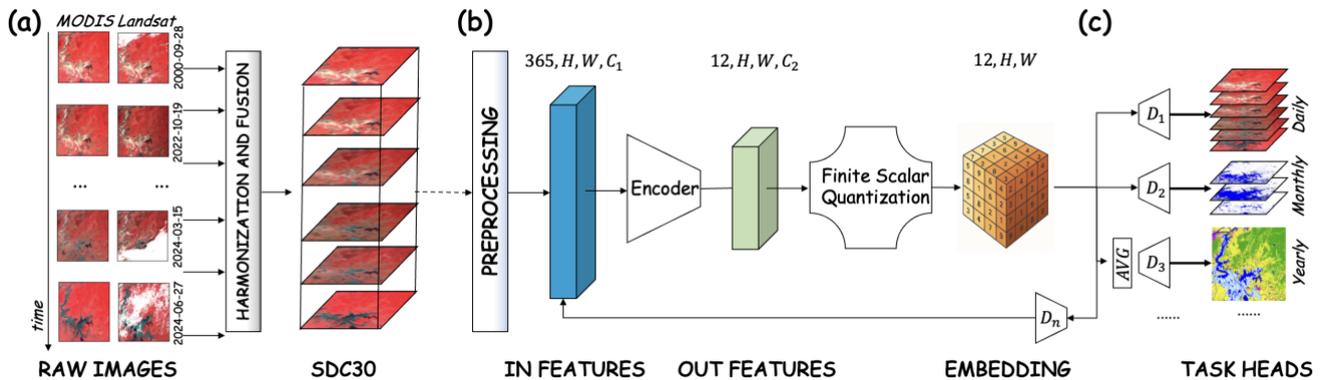

**Figure 1: Schematic of the integrated ESD production pipeline.**

## 3.2 ESDNet Architecture and Computational Framework

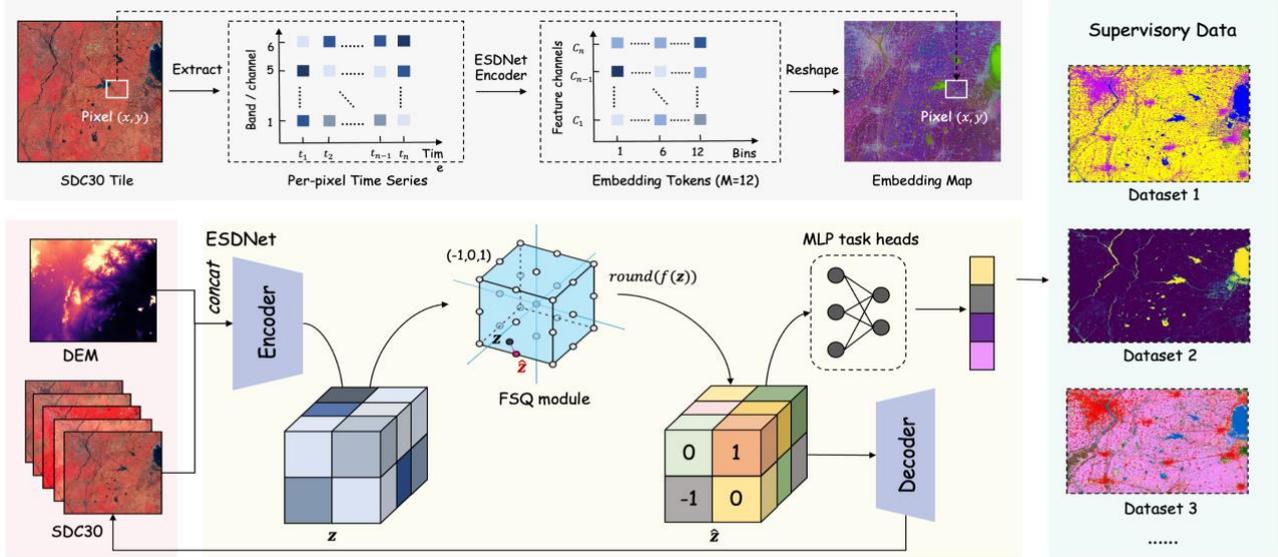

Figure 2: Detailed network architecture and computational process of ESDNet.

The ESDNet architecture is engineered to transform time-series spatiotemporal reflectance data into discrete, information-dense embeddings. As illustrated in **Figure 2**, the network employs a symmetric encoder-decoder structure integrated with a discrete quantization bottleneck and multi-task prediction heads.

**(a) Temporal Encoder Network**

The encoder is designed to extract hierarchical features from the SDC30 time-series. It consists of $N$ initial 1D convolutional (Conv1D) layers followed by $M$ residual Conv1D blocks. To reduce computational overhead and condense the temporal signal, the initial layers utilize a stride $s > 1$, performing temporal downsampling. This effectively reduces the input sequence length while broadening the receptive field to capture long-term phenological patterns. The subsequent residual blocks employ a stride of 1 and incorporate shortcut connections to facilitate the training of deeper configurations, mitigating gradient vanishing issues and ensuring the preservation of high-level semantic abstractions.

**(b) Finite Scalar Quantization (FSQ) Module**

To achieve the high compression ratios required for a global-scale database, the latent features are passed through a Finite Scalar Quantization (FSQ) module (Mentzer et al., 2023). Unlike traditional Vector Quantization (Oord et al., 2018), which relies on a learned codebook prone to "codebook collapse," FSQ projects latent representations into a bounded, lower-dimensional space where each dimension is quantized into a set of fixed, predefined scalars.

The quantization is implemented via a simple rounding operation, and gradients are propagated during backpropagation using a straight-through estimator (STE) (Bengio et al., 2013). This design creates an implicit codebook via the Cartesian product of the scalar sets, providing a highly stable and lightweight quantization scheme that eliminates the need for auxiliary commitment losses (Takida et al., 2022).

**(c) Mirror Decoder and Reconstruction**

The decoder serves as the architectural mirror of the encoder, comprising $M$ residual blocks and $N$ transposed Conv1D layers. The transposed convolutions upsample the quantized embeddings, progressively restoring the original temporal dimensions of the input SDC30 reflectance data. The primary objective of the decoder is the high-fidelity reconstruction of the input sequence; the resulting reconstruction loss provides a self-supervised signal that ensures the embeddings retain the fundamental biophysical properties of the land surface.

**(d) Multi-Task Prediction Heads**

To imbue the embeddings with explicit semantic meaning, the quantized latent vectors are processed through a temporal average-pooling layer to generate a summary feature vector. This vector is fed into multiple, independent task-specific heads—each implemented as a single-layer fully connected network. These heads are trained concurrently using the supervisory datasets described in **Section 2.2** (e.g., GAIA, ESA WorldCover, GLAD-SW, and GLAD-CE). This multi-task learning (MTL) framework constrains the latent space, forcing the embeddings to represent both spectral-temporal dynamics and high-level land-system categories.

**3.3 Multi-Task Training Strategy**

To ensure that the Embedded Seamless Data (ESD) captures both the fundamental biophysical properties and the high-level semantic categories of the land surface, we implemented a multi-task learning (MTL) framework. The model is trained using a joint objective function that combines an unsupervised reconstruction task with supervised classification and regression tasks. The total loss $L$ is defined as a weighted sum of these components:

$$L = \alpha L_{reconstruction} + \beta L_{classification} + \gamma L_{regression} \quad (1)$$

where $\alpha$, $\beta$, and $\gamma$ are hyperparameters that balance the contribution of each task to the total gradient.

**(a) Unsupervised Reconstruction Loss**

The reconstruction loss, $L_{reconstruction}$, quantifies the discrepancy between the input reflectance sequence $x$ and the reconstructed output $\hat{x}$ generated by the decoder (e.g., $D_n$ in **Figure 1**). Following a Gaussian likelihood assumption for pixel-level values, this is implemented as a Mean Squared Error (MSE) objective:

$$L_{reconstruction} = \log p(x|\hat{z}(x)) = \|x - \hat{x}\|^2 \qquad (2)$$

This self-supervised signal ensures that the quantized embeddings preserve the fine-scale spectral signatures and phenological variations necessary for near-daily and monthly surface monitoring.

**(b) Supervised Classification Loss**

The classification loss $L_{classification}$ is computed across multiple task-specific heads (eg., $D_1$, $D_2$, and $D_3$ in **Figure 1**). These heads are supervised by the diverse land-cover products described in **Section 2.2** (e.g., FROMGLC, GAIA and ESA WorldCover). For each supervisory task $i$, the loss is calculated using weighted multi-label cross-entropy:

$$L_{classification} = \sum_i a_i \sum_{k=1}^{K_i} y_k \log(\hat{y}_k) \qquad (3)$$

where $y_k$ represents the ground-truth label, $\hat{y}_k$ is the predicted probability for class $k$, and $a_i$ is a task-specific weight.

This constraint forces the latent space to cluster according to meaningful thematic categories, such as forest, water, or impervious surfaces.

**(b) Supervised Regression Loss**

To further align the embeddings with essential climate and vegetation variables, we incorporate a regression objective, $L_{regression}$. This task focuses on predicting biophysical indices derived from the SDC time series and aggregated as monthly averages, such as the Normalized Difference Vegetation Index (NDVI), Normalized Difference Water Index (NDWI), and Normalized Difference Snow Index (NDSI). The loss is formulated as:

$$L_{regression} = \sum_i b_i \|v - \hat{v}\|^2 \qquad (4)$$

where $v$ and $\hat{v}$ are the target and predicted biophysical indices, respectively. By explicitly regressing these indices, the model learns to prioritize the spectral bands and temporal windows most critical for characterizing vegetation health and water dynamics.

**3.4 Global-scale Model Training and Validation**

3.4.1 Training Samples Generation and Harmonization

To ensure robust model generalization across diverse geographic regions, climatic zones, and ecological biomes, we constructed a large-scale, globally distributed training dataset. We selected a total of 223,622 unique 6 × 6 km locations through a rigorous stratified random sampling strategy. To ensure the dataset represents the full spectrum of Earth's surface characteristics, stratification was guided by three primary dimensions: biome boundaries, continental divisions, and land-cover class proportions derived from the ESA WorldCover 2021 global map. This approach ensures adequate representation of all major thematic classes—including forests, shrublands, grasslands, croplands, urban areas, barren lands, wetlands, water bodies, and snow/ice—across varying environmental and socio-ecological contexts.

For each sampling unit, we synthesized a comprehensive suite of multi-temporal and multi-source features spanning the period from 2000 to 2024 (**Table 2**). The model's input features are categorized into two streams:

- **Dynamic Inputs**: Annual SDC30 time-series surface reflectance (30 m resolution), providing the primary spectral-temporal signal. The input shape of $[365 \times 6]$ represents the daily observations across six spectral bands, capturing the full annual phenological cycle.
- **Static Covariates**: Topographic attributes derived from NASADEM (e.g., elevation and slope), providing spatially informative context to support terrain-aware feature extraction.

To reconcile the heterogeneous temporal cadences of the supervisory labels, all products were resampled to a common 30 m grid and synchronized with the SDC30 reflectance records. As detailed in **Table 2**, we adopted a tiered synchronization logic: static products (e.g., NASADEM, WorldCover 2021) were treated as invariant across the study period; annual products (FROMGLC30, GAIA) were matched by calendar year; and composite products (e.g., GLAD-CE) were assigned to their corresponding four-year windows. For the GLAD-SW dataset, we maintained its full monthly granularity by linking each annual SDC30 observation to twelve monthly water-presence indicators, enabling the model to better characterize intra-annual hydrological variability and seasonality.

Table 2: Attributes of the multi-source satellite imagery and supervisory datasets used for ESDNet training.

| Product    | SDC30     | NASADEM | WorldCover2021 | FROMGLC30 | GAIA      | GLAD-CE   | GLAD-SW   |
|------------|-----------|---------|----------------|-----------|-----------|-----------|-----------|
| Cadence    | Yearly    | Static  | Static         | Yearly    | Yearly    | 4-Year    | Monthly   |
| Coverage   | 2000-2024 | Static  | 2021           | 2000-2024 | 1985-2024 | 2000-2019 | 1999-2024 |
| Data shape | [365×6]   | [1]     | [1]            | [1]       | [1]       | [1]       | [12]      |

The spatial distribution of these sampled locations is illustrated in **Figure 3**, colored by their dominant land-cover class. This highly diverse and balanced training foundation ensures broad spectral and spatial coverage, providing the statistical depth necessary for accurate, long-term monitoring of global land-cover dynamics.

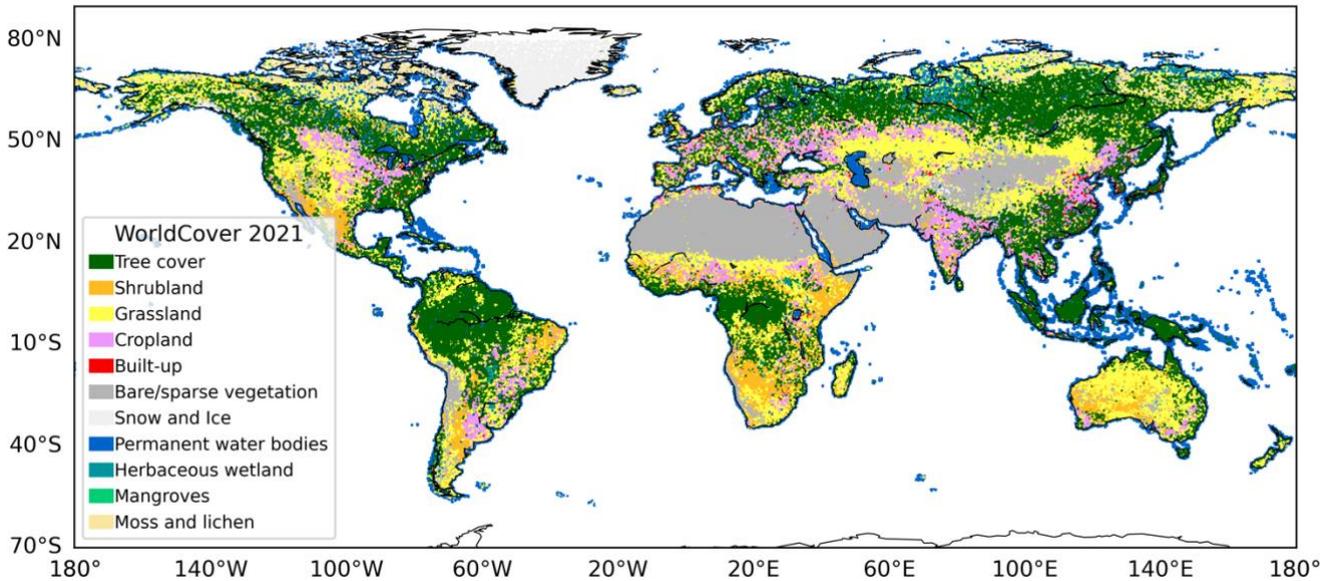

**Figure 3: Geographic distribution and thematic composition of the global training dataset.** The map displays the locations of the 223,622 stratified random sampling sites used to train the ESDNet. Each site is colored according to its dominant land-cover class as derived from the ESA WorldCover 2021 product.

3.4.2 Quantitative Evaluation: Reconstruction Fidelity of the Discrete Latent Space

To evaluate the representational capacity of the learned embeddings, we assessed the reconstructive fidelity of the ESDNet encoder-decoder architecture. This evaluation quantifies the degree to which essential spectral and temporal information is preserved within the compressed, quantized latent space. Specifically, we measured the discrepancy between the original SDC30 input reflectance ($x$) and the corresponding reconstructed output ($\hat{x}$) generated by the decoder.

Reconstruction accuracy was primarily quantified using the Mean Absolute Error (MAE), calculated independently for each of the six spectral bands across the full 365-day temporal sequence. To provide a comprehensive assessment of physical fidelity, we also computed the Root Mean Square Error (RMSE) and the Correlation Coefficient (CC). The evaluation was conducted using the FAST-validation repository, which consists of 36,636 globally distributed locations. These samples were strictly held out during the training phase to ensure an unbiased assessment of the model's generalization capabilities across diverse terrestrial biomes and phenological regimes.

3.4.3 Evaluation of Generalization and Transferability to Unseen Tasks

A core objective of the ESDNet is to provide "universal" representations that support generalization beyond the specific supervised tasks encountered during training. To test this capability, we conducted two transfer learning experiments where the learned embeddings were utilized as fixed, frozen feature representations for novel classification problems.

**In-Domain Transfer**: We trained a diverse set of downstream classifiers—including linear prediction heads, Ridge Classifiers, K-Nearest Neighbors (KNN), and Random Forests—using the FAST-training set and evaluated their performance on the FAST-validation set.

**Out-of-Domain Generalization**: To assess the model's robustness against external benchmarks, we utilized the official training and validation samples provided by the Copernicus Global Land Operations Service (CGLOPS).

For both experiments, we employed the same suite of performance metrics (OA, precision, recall, and F1-score). Furthermore, we analyzed model performance under varying data regimes to evaluate the few-shot learning capabilities of the ESD embeddings. By measuring accuracy across a range of training sample sizes, this analysis provides critical insights into the practical utility of the representations in data-scarce settings and their adaptability to emergent downstream applications in Earth system science.

## 4. Results and Analysis

### 4.1 Global Product Characteristics and Data Organization

The Embedded Seamless Data (ESD) provides a high-fidelity, continuous 30 m resolution representation of the global land surface, spanning a 25-year longitudinal record from 2000 to 2024. By leveraging the robust reconstructive capabilities of the ESDNet decoder and the initial spatiotemporal fusion of the SDC30 foundation, the resulting embeddings effectively eliminate the traditional barriers to global-scale analysis, such as "seam-line" artifacts, cloud-contamination, and sensor-specific biases.

As illustrated in **Figure 4**, the ESD maintains remarkable spatial and spectral consistency across continental scales. The transition between different Landsat generations (5, 7, 8, and 9) is handled seamlessly, providing the longitudinal stability required for decadal Earth system studies. Despite the high compression ratio, the embeddings preserve fine-grained landscape features—including intricate river networks, urban morphological boundaries, and fragmented agricultural patterns—ensuring that the latent space acts as a reliable proxy for raw surface reflectance.

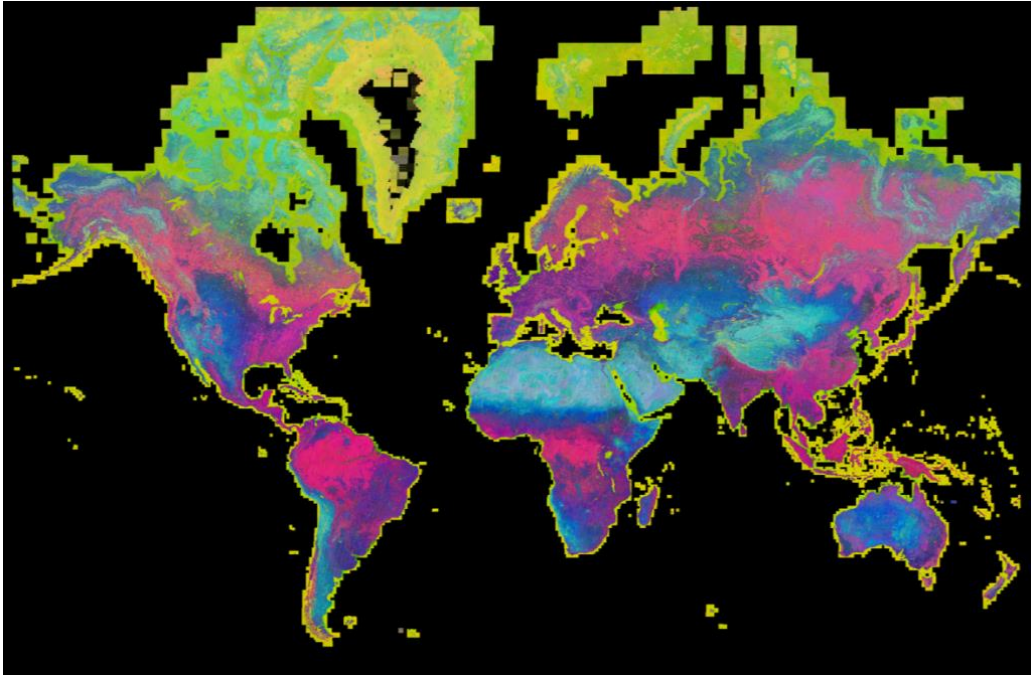

**Figure 4: Global visualization of the 30-m Embedded Seamless Data (ESD) product for the year 2024.**

To ensure maximum interoperability with existing Earth Observation (EO) workflows, the ESD is organized according to the Military Grid Reference System (MGRS) in the Universal Transverse Mercator (UTM) projection, matching the tiling convention of the Sentinel-2 mission (Drusch et al., 2012).

- **Dimensions**: Every tile follows a fixed structure of [12, 3600, 3600]. The first dimension ($12$) represents 12 temporal latent steps, which correspond to monthly information-dense summaries that capture intra-annual phenological dynamics. The remaining dimensions (3600×3600) correspond to the 30-m spatial grid, covering a standard 108×108 km area.
- **Encoding and Quantization:** The embeddings are stored as quantized integers derived from the Finite Scalar Quantization (FSQ) module. By utilizing integer-based encoding, the file footprint is significantly reduced without a concomitant loss in semantic or reconstructive accuracy.

The most transformative characteristic of the ESD product is its unprecedented efficiency in data volume management. As shown in **Table 3**, a single year of global, daily 30-m surface reflectance in raw formats would exceed several petabytes—a scale that precludes analysis by most research teams. In contrast, the ESD encapsulates the entire global land surface for one year within approximately 2.4 TB. This ~340-fold reduction in volume allows researchers to host and process decadal-scale global datasets on standard local workstations or modest server clusters, removing the financial and technical barriers associated with petabyte-scale cloud computing. Consequently, the ESD facilitates a new era of agile Earth system research where global-scale analyses can be executed with the same ease as local-scale studies.

Table 3: Comparison of data volumes between the SDC30 and the ESD product.

| Data volume for one year | Each tile (108 × 108 km) | China mainland (1224 tiles) | Global land (18466 tiles) |
|---|---|---|---|
| SDC30 | 45.6 GB (128MB * 365) | 54.5 TB | 0.8 PB |
| ESD | 136.3 MB | 162.9 GB | 2.4 TB |

## 4.2 Evaluation of Reconstructive Accuracy

The primary technical benchmark for the ESD is its ability to preserve the biophysical signal of the original 30 m surface reflectance despite the high compression ratios achieved. This fidelity ensures that the information-dense latent representations can serve as a reliable, physically meaningful proxy for raw satellite observations in downstream Earth system research.

We evaluated the reconstruction accuracy using the FAST-validation set, a globally distributed repository comprising 36,636 independent locations held out during the model training phase. Accuracy was quantified across the six primary spectral bands—Blue, Green, Red, NIR, SWIR1, and SWIR2—using Mean Absolute Error (MAE), Root Mean Square Error (RMSE), and the Correlation Coefficient (CC).

The results, summarized in **Table 4**, demonstrate that the ESDNet encoder-decoder architecture maintains high reconstructive fidelity across the entire solar reflective spectrum. The mean MAE across all bands is approximately 0.0130, with a mean RMSE of 0.0179. The highest reconstructive accuracy was observed in the SWIR2 spectral region (MAE = 0.0115), whereas the NIR band exhibited the highest absolute error (MAE = 0.0170). This variation is consistent with the higher dynamic range and inherent phenological variability typically found in near-infrared observations of vegetated surfaces. The mean Correlation Coefficient (CC) remains high at 0.854. This confirms that the Finite Scalar Quantization (FSQ) bottleneck effectively retains the spectral-temporal signatures necessary for resolving fine-grained land surface dynamics.

Table 4: Decoder reconstruction accuracy in MAE, RMSE, and CC evaluated using the FAST-validation set.

| Band | Blue | Green | Red | NIR | SWIR1 | SWIR2 | Mean |
|---|---|---|---|---|---|---|---|
| MAE↓ | 0.0121 | 0.0116 | 0.0123 | 0.0170 | 0.0139 | 0.0115 | 0.0130 |
| RMSE↓ | 0.0176 | 0.0166 | 0.0175 | 0.0227 | 0.0183 | 0.0150 | 0.0179 |
| CC↑ | 0.8023 | 0.8494 | 0.8801 | 0.8848 | 0.8681 | 0.8411 | 0.8543 |

A critical requirement for Earth System Science Data is the stability of the signal over time. Temporal analysis illustrated in **Figure 5** reveals that the reconstructive fidelity is remarkably consistent throughout the 25-year study period.

The stability of the MAE and CC metrics over two decades demonstrates the model's robustness against transitions between underlying satellite constellations, specifically from Landsat-5 TM and 7 ETM+ to the newer Landsat-8/9 OLI sensors. This longitudinal consistency ensures that the ESD can be used for tracking decadal shifts in Earth system processes—such as forest degradation, urban expansion, and water body fluctuations—without introducing artificial "jumps" caused by sensor changes.

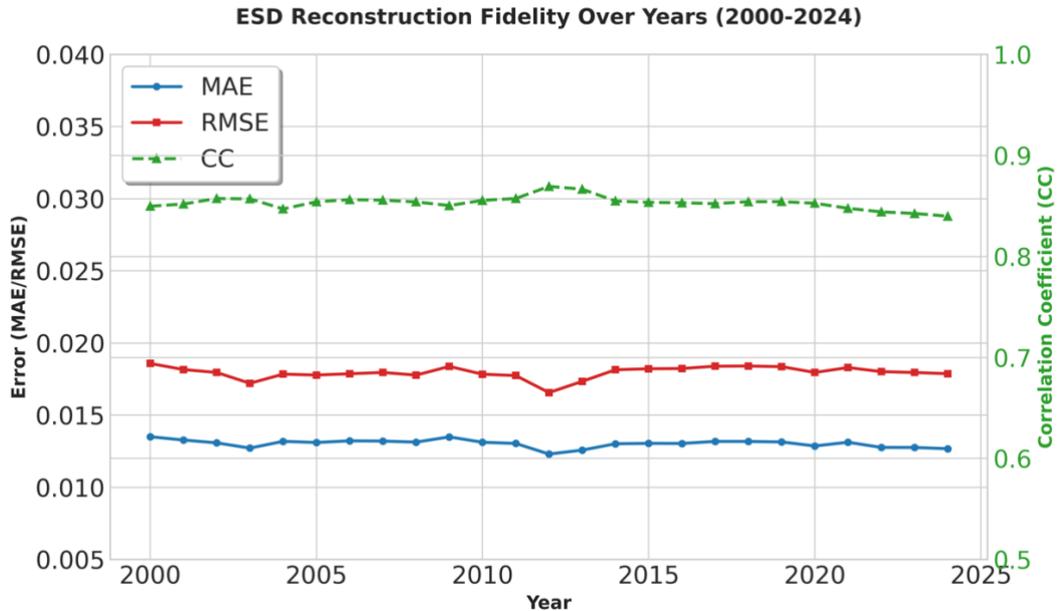

**Figure 5: Temporal evaluation of the reconstructive fidelity of the ESDNet encoder-decoder architecture (2000–2024).** The plot illustrates the Mean Absolute Error (MAE), Root Mean Square Error (RMSE), and Correlation Coefficient (CC) calculated between the input SDC30 surface reflectance and the reconstructed output generated from the quantized latent space. Metrics represent the average performance across all six spectral bands (Blue, Green, Red, NIR, SWIR1, and SWIR2). Validation was performed using the globally distributed FAST-validation repository, comprising independent samples held out during the model training phase.

Beyond statistical metrics, the ESDNet's ability to preserve the structural integrity of the landscape is vital. As shown in the detailed tile evaluations (**Figures 7–10**), the model effectively reconstructs essential landscape features even in regions with challenging conditions: a) Intricate river networks, crisp urban boundaries, and individual agricultural field patterns are clearly maintained in the reconstructed output; b) The reconstruction process demonstrates a notable "denoising" effect, where residual artifacts from cloud-contamination or topography in the original SDC30 data are minimized in the latent representation; c) The spatial distribution of MAE, visualized **in Figure 6**, confirms that low error rates are achieved globally, across diverse terrestrial biomes and extreme topographic reliefs.

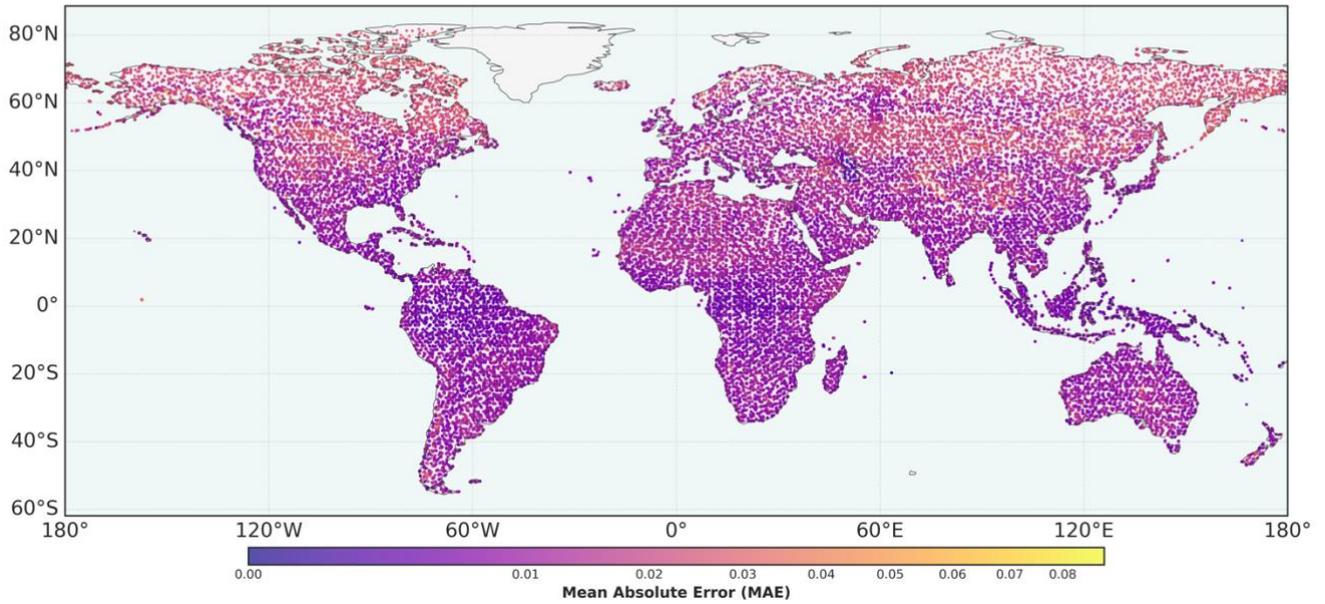

**Figure 6**: Global spatial distribution of reconstructive fidelity in MAE for the Embedded Seamless Data (ESD) product.

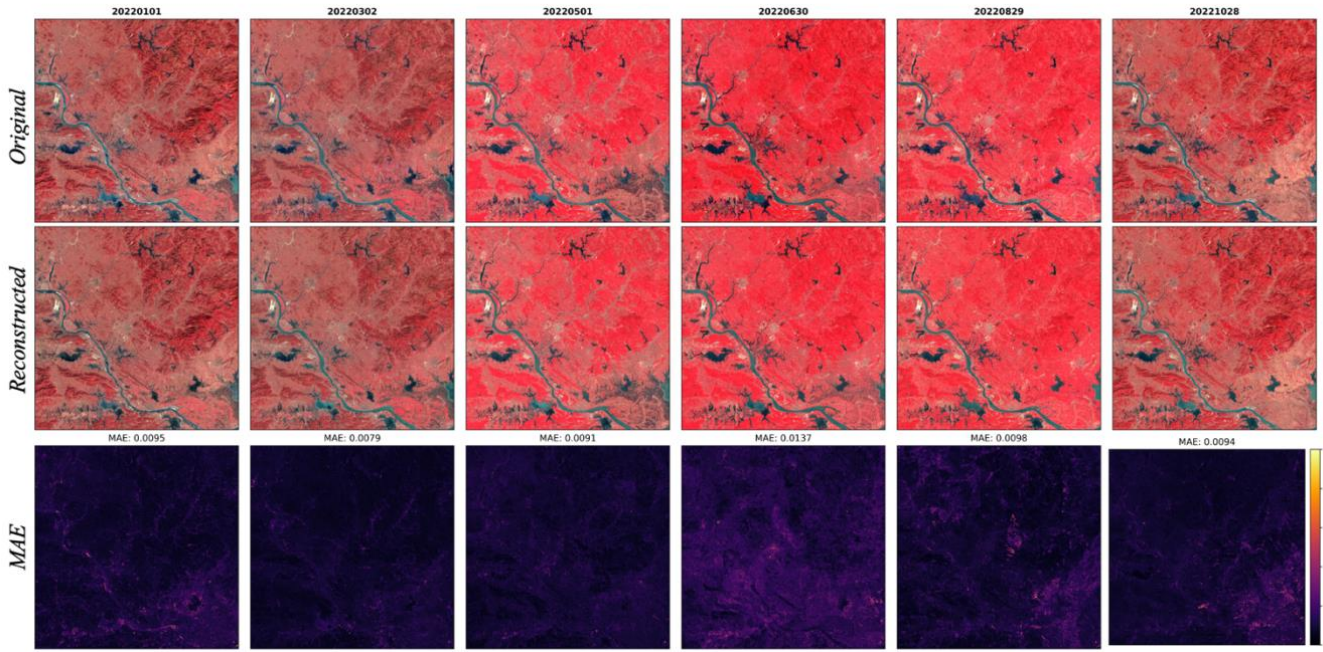

**Figure 7**: Reconstructive performance in tile 50RLU. The top row displays the original SDC30 surface reflectance (NIR-Red-Green false-color composite). The middle row shows the corresponding reconstructed output from the ESDNet decoder. The bottom row quantifies the spatial distribution of the MAE per pixel.

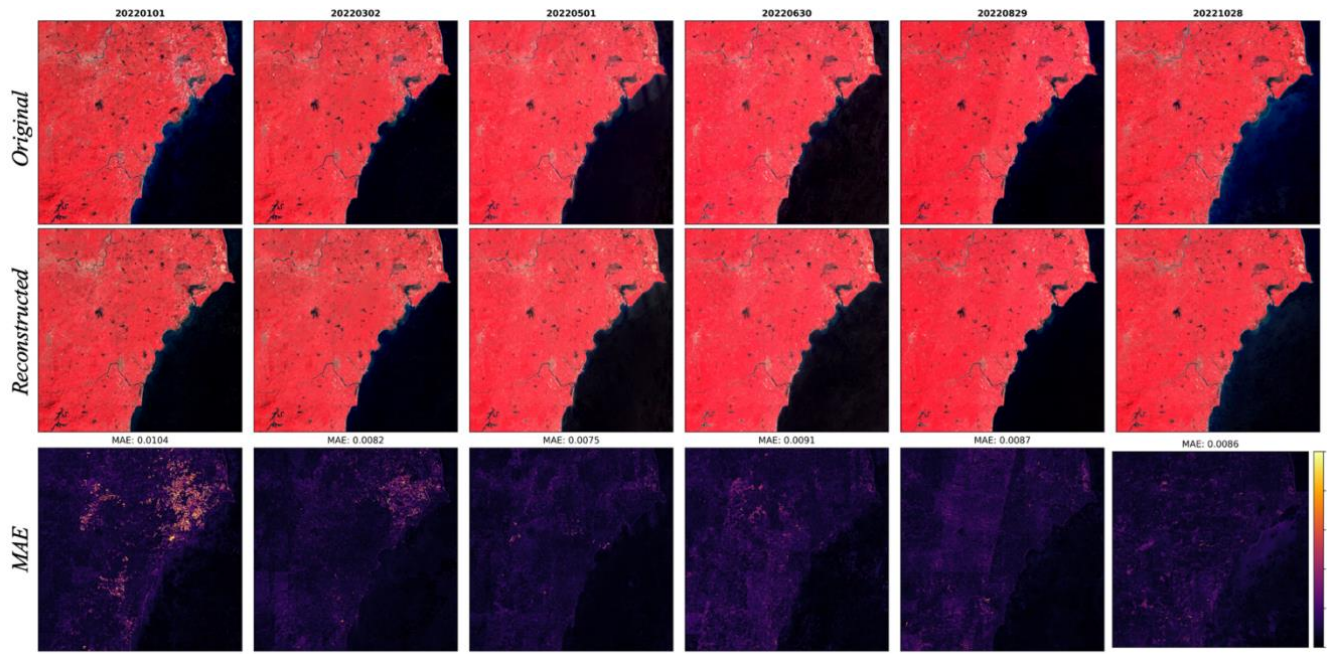

**Figure 8:** Reconstructive performance in tile 49QDB. The top row displays the original SDC30 surface reflectance (NIR-Red-Green false-color composite). The middle row shows the corresponding reconstructed output from the ESDNet decoder. The bottom row quantifies the spatial distribution of the MAE per pixel.

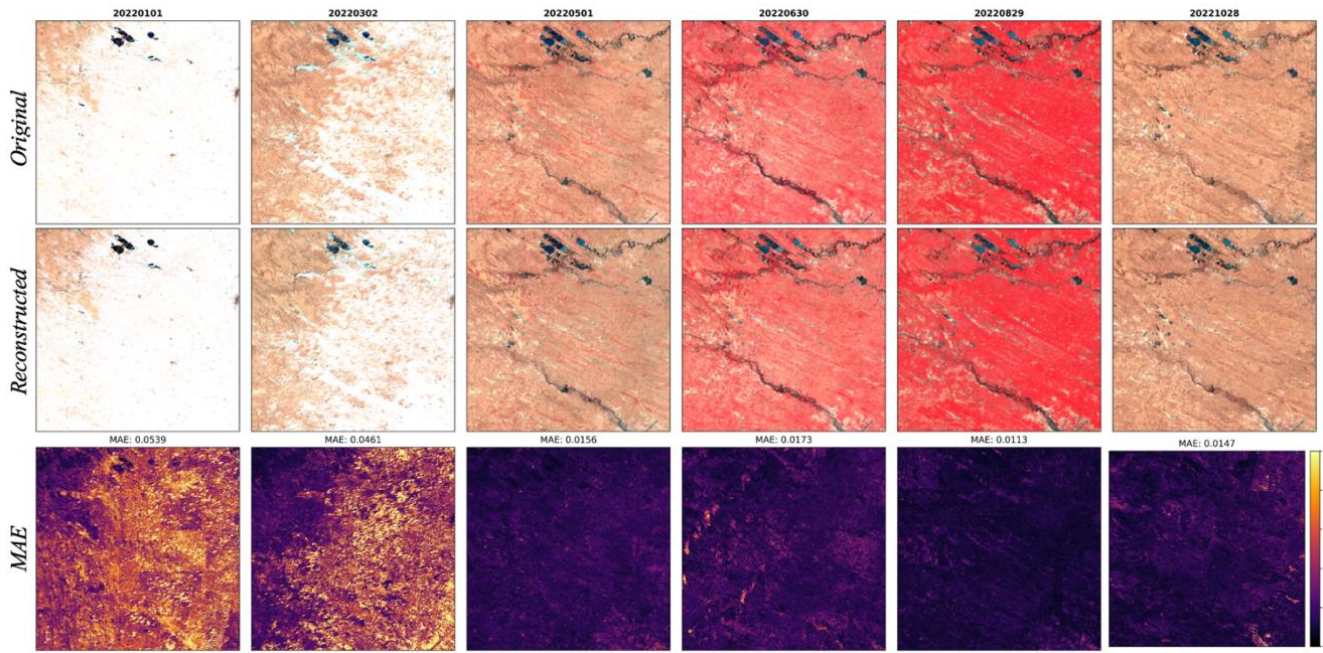

**Figure 9:** Reconstructive performance in tile 51TVK. The top row displays the original SDC30 surface reflectance (NIR-Red-Green false-color composite). The middle row shows the corresponding reconstructed output from the ESDNet decoder. The bottom row quantifies the spatial distribution of the MAE per pixel.

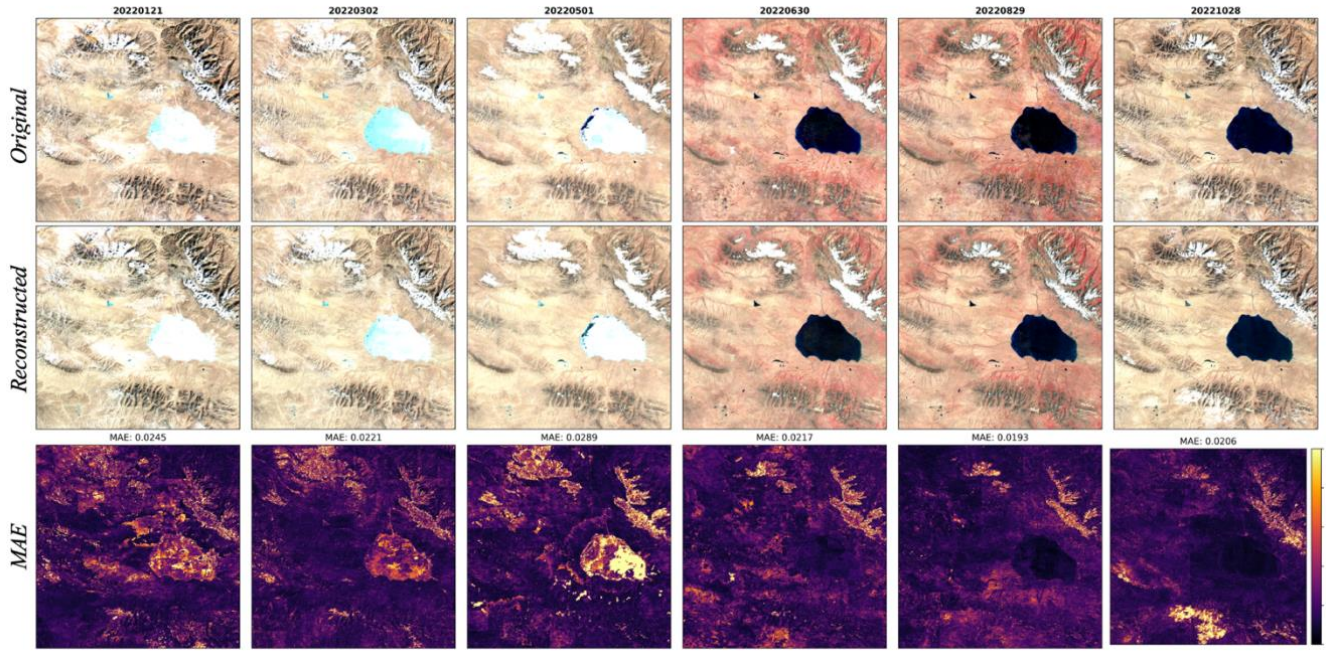

**Figure 10: Reconstructive performance in tile 45SXB.** The top row displays the original SDC30 surface reflectance (NIR-Red-Green false-color composite). The middle row shows the corresponding reconstructed output from the ESDNet decoder. The bottom row quantifies the spatial distribution of the MAE per pixel.

**4.3 Transferability and Generalization Analysis**

A primary objective of the Embedded Seamless Data (ESD) product is to provide "universal" representations that support robust generalization beyond the specific supervised tasks encountered during its construction. To evaluate this, we benchmarked the performance of the ESD latent vectors against the original high-dimensional SDC30 reflectance data and standard temporal compositing methods across various downstream classification tasks.

4.3.1 Downstream Classification Performance

We conducted an in-domain transfer experiment using four distinct thematic monitoring tasks: FROMGLC, ESA WorldCover, GLAD Global Crop Extent, and GLAD Global Surface Water. To compare the thematic accuracy of classifications derived from ESD versus raw SDC30 data and Landsat compositing, we employed a suite of common machine learning algorithms, including Linear Prediction Heads, k-Nearest Neighbors (k=1 and k=3), and Random Forests.

As illustrated in **Figure 11**, using ESD achieves consistently higher performance across all tested algorithms and tasks. Despite a significant reduction in data dimensionality, the ESD latent vectors effectively distill and preserve essential semantic features. A detailed comparison of the confusion matrices for land-cover classification using the Random Forest algorithm is provided in **Tables 5 and 6**. ESD achieved a higher global OA of 79.74% compared to 76.92% for the raw SDC30 data. Significant gains were observed in several primary categories. For instance, the Producer's Accuracy (PA) for Crop improved from 61.82% (SDC30) to 71.96% (ESD), while the PA for Grass rose from 64.01% to 70.25%. The User's Accuracy (UA) for Impervious surfaces saw a notable increase from 84.38% to 80.62%, and Water reached a high of 95.29%.

These results indicate that the ESD latent space is better organized to resolve spectral-temporal ambiguities, thereby enhancing the classification of complex land cover types.

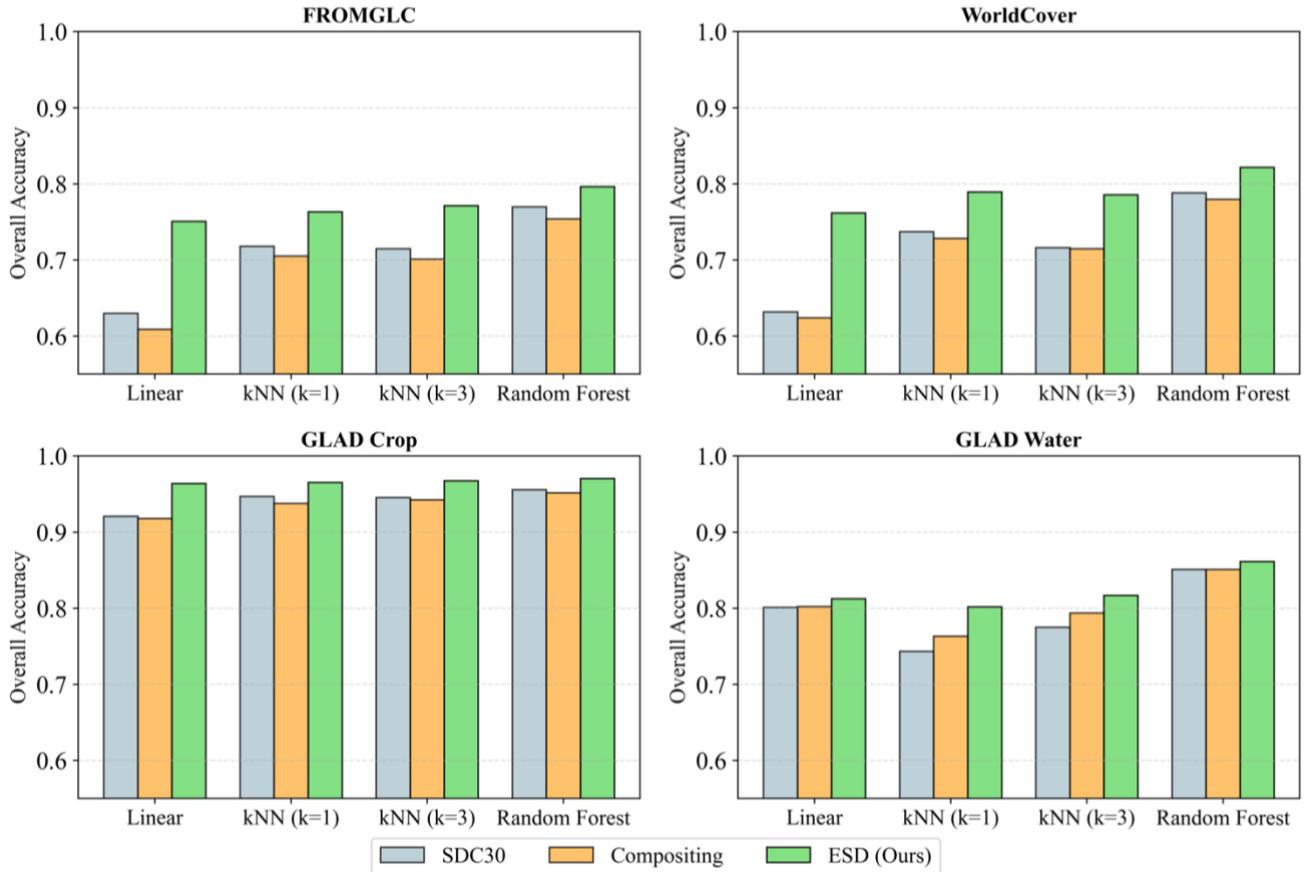

**Figure 11:** Comparison of downstream classification performance across diverse thematic monitoring tasks. The bar charts illustrate the Overall Accuracy (OA) achieved using the Embedded Seamless Data (ESD) latent representations, the Global 30 m Seamless Data Cube (SDC30) reflectance, and temporal compositing methods. Evaluation is conducted across four distinct tasks: FROMGLC , ESA WorldCover , GLAD Global Crop Extent , and GLAD Global Surface Water. For each task, performance is benchmarked using four classification algorithms: Linear Prediction Head, k-Nearest Neighbors (k=1 and k=3), and Random Forest.

**Table 5:** Confusion matrix of land-cover classification results using the Global 30 m Seamless Data Cube (SDC30) and Random Forest. The classification was performed on the independent FAST-validation sample set. Producer's Accuracy (PA) and User's Accuracy (UA) are provided for each of the primary land-cover categories.

| Class | Crop | Forest | Grass | Shrub | Water | Tundra | Impervious | Bareland | Snow/Ice | PA |
|---|---|---|---|---|---|---|---|---|---|---|
| Crop | 2048 | 154 | 808 | 247 | 3 | 0 | 2 | 51 | 0 | 61.82% |
| Forest | 116 | 8793 | 449 | 320 | 7 | 78 | 0 | 4 | 0 | 90.03% |
| Grass | 362 | 596 | 4247 | 780 | 33 | 225 | 5 | 387 | 0 | 64.01% |
| Shrub | 176 | 676 | 907 | 2669 | 5 | 58 | 3 | 185 | 0 | 57.04% |
| Water | 3 | 16 | 24 | 3 | 1340 | 22 | 0 | 24 | 0 | 93.58% |
| Tundra | 6 | 69 | 165 | 3 | 18 | 1196 | 0 | 71 | 0 | 78.27% |
| Impervious | 57 | 18 | 75 | 25 | 1 | 0 | 54 | 42 | 0 | 19.85% |
| Bareland | 5 | 7 | 203 | 108 | 24 | 50 | 0 | 5221 | 7 | 92.82% |
| Snow/Ice | 0 | 0 | 1 | 0 | 5 | 0 | 0 | 20 | 118 | 81.94% |
| UA | 73.86% | 85.13% | 61.74% | 64.24% | 93.31% | 73.42% | 84.38% | 86.94% | 94.40% | **76.92%** |

Table 6: Confusion matrix of land-cover classification results using the Embedded Seamless Data (ESD) and Random Forest. The classification was performed on the independent FAST-validation sample set. Producer's Accuracy (PA) and User's Accuracy (UA) are provided for each of the primary land-cover categories.

| Class | Crop | Forest | Grass | Shrub | Water | Tundra | Impervious | Bareland | Snow/Ice | PA |
|---|---|---|---|---|---|---|---|---|---|---|
| Crop | 2384 | 136 | 609 | 141 | 3 | 0 | 9 | 31 | 0 | 71.96% |
| Forest | 72 | 8902 | 412 | 319 | 7 | 46 | 5 | 4 | 0 | 91.14% |
| Grass | 322 | 515 | 4661 | 592 | 25 | 196 | 8 | 316 | 0 | 70.25% |
| Shrub | 96 | 683 | 923 | 2766 | 2 | 49 | 1 | 159 | 0 | 59.12% |
| Water | 3 | 15 | 35 | 1 | 1334 | 22 | 1 | 21 | 0 | 93.16% |
| Tundra | 0 | 65 | 208 | 4 | 12 | 1159 | 0 | 80 | 0 | 75.85% |
| Impervious | 24 | 24 | 70 | 19 | 2 | 0 | 104 | 29 | 0 | 38.24% |
| Bareland | 7 | 7 | 242 | 91 | 12 | 49 | 1 | 5210 | 6 | 92.62% |
| Snow/Ice | 0 | 0 | 0 | 0 | 3 | 0 | 0 | 31 | 110 | 76.39% |
| UA | 81.98% | 86.03% | 65.10% | 70.33% | 95.29% | 76.20% | 80.62% | 88.59% | 94.83% | **79.74%** |

4.3.2 Few-Shot Learning Capability

Sample size is important in global land cover mapping (Gong et al., 2024, 2019). To assess the practical utility of ESD in data-scarce environments—a common challenge in global Earth Observation—we conducted a few-shot learning examination on the FAST-validation set. We compared the performance of ESD and SDC30 across a range of training sample sizes, from $10^2$ to nearly $10^5$ samples.

As illustrated in **Figure 12**, ESD consistently outperforms SDC30 in data-scarce regimes. Specifically, the ESD-based linear classifier reaches a performance plateau much earlier in the sample size progression compared to SDC30, which struggles to maintain accuracy at lower sample counts. Even with only 100 training samples, ESD-based models provide significantly higher Overall Accuracy than their raw-data counterparts. This demonstrates that the ESD latent space is already pre-organized into meaningful semantic clusters. This few-shot advantage is maintained across multiple algorithms, including KNN and Random Forest, highlighting the "universal" nature of the learned representations and their adaptability to emergent downstream applications in Earth system science.

By reducing the requirement for massive labeled datasets, the ESD product enables more agile and cost-effective environmental monitoring, particularly for regions or phenomena where ground-truth data is difficult to obtain

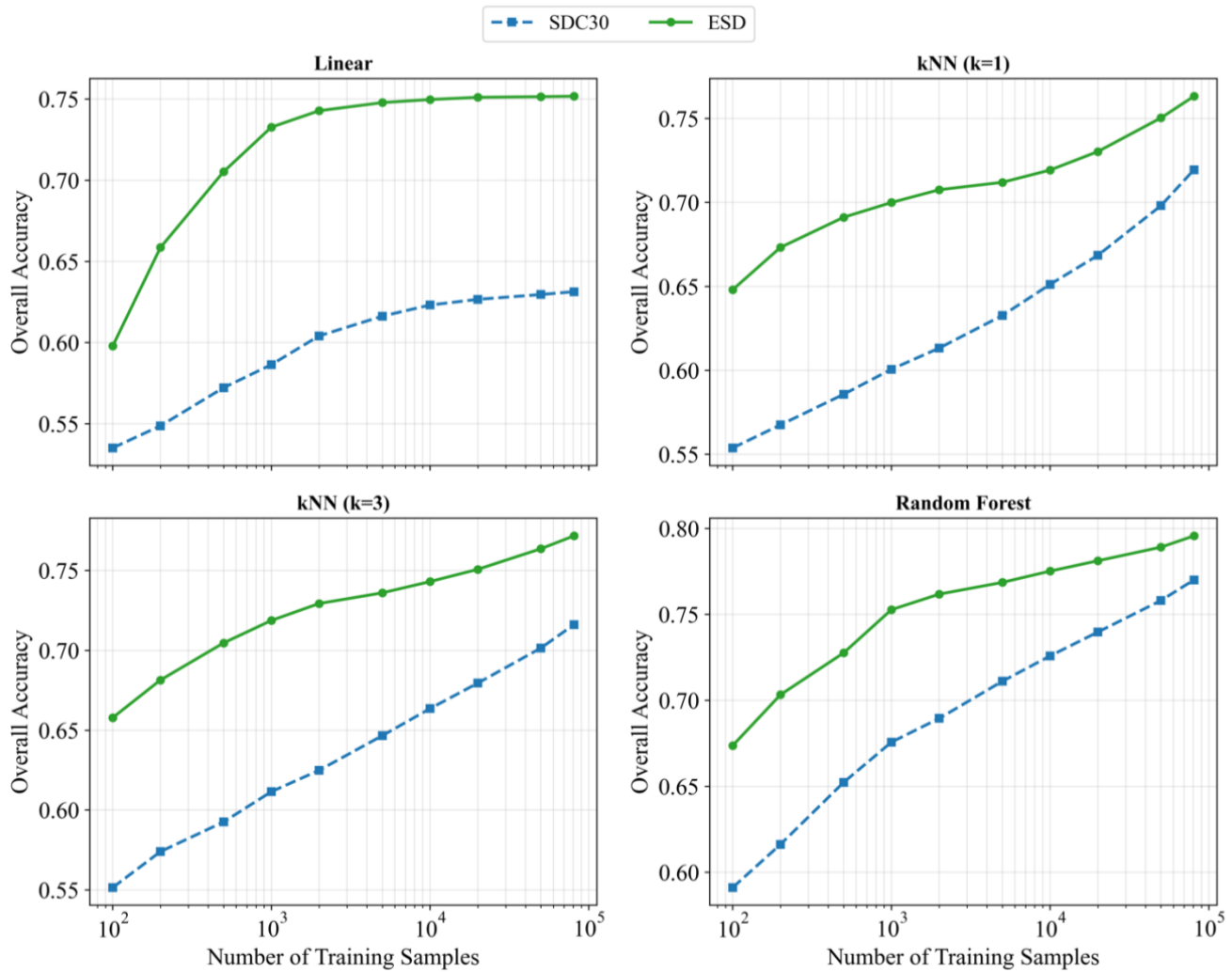

Figure 12: Comparison of downstream classification performance across diverse thematic monitoring tasks. The bar charts illustrate the Overall Accuracy (OA) achieved using the Embedded Seamless Data (ESD) latent representations, the Global 30 m Seamless Data Cube (SDC30) reflectance, and temporal compositing methods. Evaluation is conducted across four distinct tasks: FROMGLC ,

ESA WorldCover , GLAD Global Crop Extent , and GLAD Global Surface Water. For each task, performance is benchmarked using four classification algorithms: Linear Prediction Head, k-Nearest Neighbors (k=1 and k=3), and Random Forest.

## 5. Discussion

### 5.1 Implicit Denoising Effects via the Encoder-Decoder Process

A significant emergent property of the ESDNet architecture is its inherent capacity to "denoise" satellite time-series data during the latent encoding and reconstruction stages . While the primary goal of the framework is dimensionality reduction and feature extraction, the process of bottlenecking high-dimensional observations into a compact, quantized latent space effectively filters out transient non-geophysical noise. By projecting raw spectral data into a unified, lower-dimensional latent space, the model is forced to prioritize the persistent biophysical signals—such as vegetation phenology and underlying surface reflectance—over sporadic atmospheric interference. This phenomenon aligns with established deep learning principles where a constrained bottleneck prevents the network from memorizing stochastic noise, instead compelling it to learn the most salient and stable hierarchical patterns of the terrestrial surface.

The practical utility of this denoising capability is clearly illustrated in the various case studies provided in **Figure 13**. In the first three columns of the figure, raw SDC30 images contain persistent cloud residuals—common artifacts that often survive traditional cloud-masking pipelines—which are effectively mitigated in the images decoded from the ESD latent space. The reconstruction process leverages the temporal depth of the embeddings to "fill in" these corrupted pixels with physically consistent data derived from the annual phenological cycle. Similarly, the final two columns of **Figure 13** highlight instances where cloud shadows in the raw SDC30 data were successfully compressed during the ESDNet decoding process.

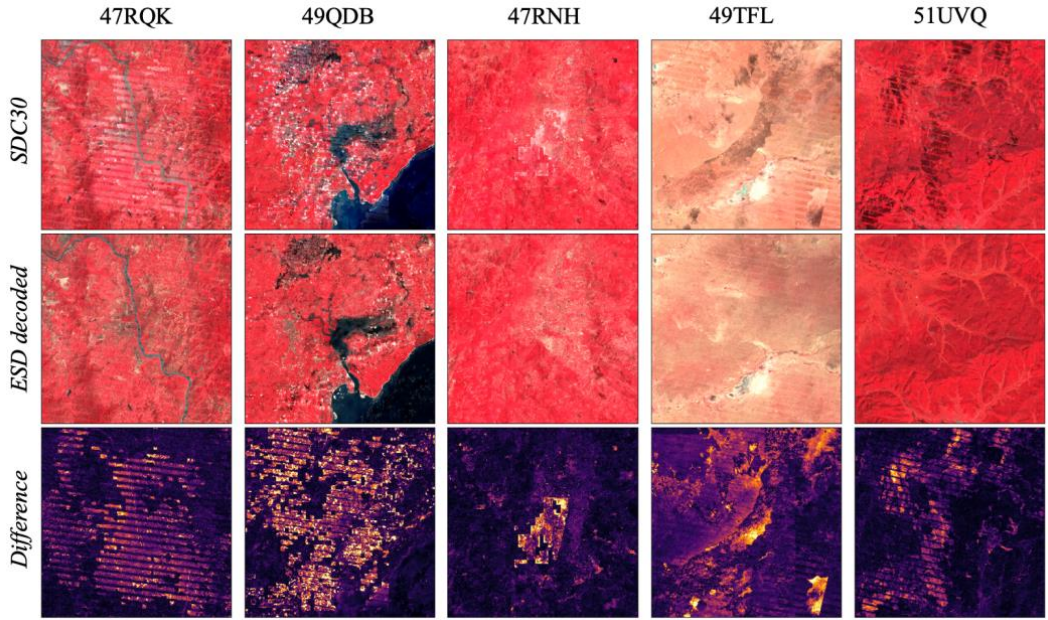

Figure 13: Examples of ESD denoising effects. The first three columns show examples that the cloud residuals in raw SDC30 were effectively mitigated in the images decoded from the ESD. The last two columns highlight instances where cloud shadows in the raw SDC30 data were successfully compressed.

## 5.2 Ablation study

To determine the optimal configuration for the Embedded Seamless Data (ESD) product, we conducted a systematic ablation study focusing on the core architectural hyperparameters of ESDNet. These experiments evaluate the trade-offs between reconstructive fidelity (measured by MAE, RMSE, and CC) and downstream semantic performance (Overall Accuracy on the ESA WorldCover task using a linear head), while considering the practical constraints of global-scale data storage.

The temporal dimension of the embeddings represents the number of latent steps used to summarize the annual phenological cycle. As shown in **Table 7**, increasing the temporal steps from 4 to 24 results in a monotonic improvement in reconstruction accuracy (MAE decreases from 0.0178 to 0.0101). However, higher temporal dimensionality directly increases the storage footprint of the database. We selected a temporal dimension of 12 as the final configuration. This choice serves two primary functions: a) It achieves high reconstructive fidelity (MAE = 0.0130) and competitive downstream performance (76.4%) while maintaining the ultra-lightweight 2.4 TB/year target; b) It naturally aligns the latent steps with the 12-month seasonal cycle, providing an intuitive structure for researchers analyzing intra-annual land surface dynamics.

Table 7: Performance comparison of varying temporal dimensions for the Embedded Seamless Data (ESD) product. Metrics include reconstruction accuracy (MAE, RMSE, CC) and downstream classification accuracy (OA) for the ESA WorldCover task. The asterisk (*) denotes the parameter (12) selected for the final product to align with the annual monthly cycle.

| Temporal dimension | MAE↓ | RMSE↓ | CC↑ | OA↑ (WorldCover, Linear) |
| --- | --- | --- | --- | --- |
| 4 | 0.0178 | 0.0246 | 0.7652 | 75.5% |
| 6 | 0.0162 | 0.0224 | 0.7937 | 75.3% |
| 8 | 0.0145 | 0.0200 | 0.8309 | 76.2% |
| **12*** | 0.0130 | 0.0179 | 0.8537 | 76.4% |
| 24 | 0.0101 | 0.0138 | 0.9035 | 76.8% |

The size of the discrete latent space (the virtual codebook) determines the "vocabulary" available to the model to represent Earth's diversity. Our results in Table 8 demonstrate that expanding the embedding dimension from 256 to 65,536 significantly enhances both spectral reconstruction and land-cover classification accuracy. We finalized the embedding dimension at 65,536 ($2^{16}$). This specific value is strategically chosen to allow each embedding value to be stored in a single uint16 format. This hardware-efficient encoding minimizes storage overhead without the semantic degradation observed at lower bit-depths, ensuring the "democratization" of the data for users with limited computational resources.

Table 8: Performance metrics across different Finite Scalar Quantization (FSQ) embedding dimensions. The asterisk (*) indicates the dimension (65,536) used for the final ESD product, chosen to balance representational accuracy with efficient uint16 data storage.

| Embedding dimension | MAE↓ | RMSE↓ | CC↑ | OA↑ (WorldCover, Linear) |
|---|---|---|---|---|
| 256 | 0.0159 | 0.0215 | 0.8056 | 71.4% |
| 1024 | 0.0145 | 0.0198 | 0.8271 | 74.3% |
| 4096 | 0.0142 | 0.0193 | 0.8364 | 74.6% |
| 16384 | 0.0136 | 0.0187 | 0.8486 | 75.1% |
| **65536*** | 0.0131 | 0.0180 | 0.8543 | 77.3% |

The complexity of feature extraction was evaluated by varying the number of residual layers within the encoder and decoder. As indicated in Table 9, performance initially improves with depth but begins to plateau and eventually decline after 10 layers. The 10-layer configuration achieved the highest downstream accuracy (77.3%) while keeping reconstruction errors low. Increasing the depth to 20 layers resulted in a slight decrease in accuracy (76.9%), likely due to diminishing returns in feature abstraction or increased optimization complexity. Consequently, 10 residual layers were selected to maximize performance efficiency.

Table 9: Performance metrics across different number of residual Conv1D layers. The asterisk (*) highlights the 10-layer configuration chosen for the final product to optimize the trade-off between model accuracy and computational overhead.

| Num of residual layers | MAE↓ | RMSE↓ | CC↑ | OA↑ (WorldCover, Linear) |
|---|---|---|---|---|
| 0 | 0.0146 | 0.0198 | 0.8348 | 72.0% |
| 1 | 0.0132 | 0.0181 | 0.8538 | 75.8% |
| 2 | 0.0126 | 0.0173 | 0.8641 | 75.5% |
| 5 | 0.0127 | 0.0174 | 0.8633 | 75.2% |
| **10*** | 0.0131 | 0.0180 | 0.8543 | 77.3% |
| 20 | 0.0133 | 0.0183 | 0.8497 | 76.9% |

A pivotal finding of our ablation study is the impact of supervised signals during training. As shown in **Table 10**, an unsupervised model (trained solely on reconstruction) achieves superior reconstruction metrics (MAE = 0.0073) but fails significantly in downstream classification (OA = 60.8%). In contrast, the multi-task supervision framework intentionally introduces a constraint on the latent space. While this slightly increases reconstruction error (MAE = 0.0131), it boosts downstream accuracy to 76.2%. This trade-off confirms that our multi-task strategy is essential for imbuing the ESD embeddings with the semantic richness required for "universal" representation learning, moving beyond simple data compression.

Table 10: Performance metrics with and without multi-task supervisory training.

| Supervisory training | MAE↓ | RMSE↓ | CC↑ | OA↑ (WorldCover, Linear) |
|---|---|---|---|---|
| Without | 0.0073 | 0.0103 | 0.9389 | 60.8% |
| **With*** | 0.0131 | 0.0180 | 0.8543 | 76.2% |

**5.3 Limitations and Future Directions**

While the Embedded Seamless Data (ESD) dataset provides a high-fidelity and ultra-lightweight framework for global Earth monitoring, several inherent limitations remain that define the scope of its current application and suggest pathways for future development.

**Current Constraints in Thematic Representation:** A primary limitation involves the trade-off between reconstruction fidelity and semantic richness. As demonstrated in our ablation studies, incorporating supervisory signals for multi-task training slightly increases the reconstruction error compared to a purely self-supervised model. This suggests that the latent space must balance "memorizing" raw spectral-temporal signatures with "clustering" according to high-level thematic categories like those found in the WorldCover or FROMGLC datasets. Furthermore, while the current embeddings effectively capture the 12-month phenological cycle, they may struggle to represent rapid, non-cyclical events—such as sudden flash floods or abrupt wildfire disturbances—that occur at a temporal scale finer than the current latent step resolution.

**Potential for Multi-Modal Integration:** The current architecture is primarily optimized for optical and topographic data from the Landsat, MODIS, and NASADEM archives. A promising future direction involves the integration of multi-modal observations, particularly Synthetic Aperture Radar (SAR) data from missions like Sentinel-1. Incorporating SAR would enhance the dataset's utility in regions with persistent cloud cover and provide structural information that complements the spectral-temporal features currently captured by the ESDNet encoder.

**Scaling Toward Higher Spatial Resolution:** Expanding the ESD framework to 10-m resolution by integrating the Sentinel-2 constellation represents another critical frontier. While the current 30-m resolution is sufficient for decadal-scale global studies, 10-m embeddings would significantly improve the characterization of fragmented landscapes, small-scale agriculture, and complex urban morphologies. Such an expansion would require addressing the substantial increase in data volume, potentially through more advanced quantization schemes or hierarchical embedding strategies.

**Adaptive and Foundation Model Interoperability:** The static nature of the current quantization codebook, while stable, limits the model's ability to adapt to entirely new sensor types without retraining. Future iterations could explore the use of dynamic codebooks or cross-modal foundation models that allow the ESD to serve as a universal "plug-and-play" latent representation for an even broader range of downstream Earth system science tasks. By continuing to refine these information-dense vectors, we aim to further lower the barriers to entry for planetary-scale environmental analysis.

## 6. Code and data availability

The Embedded Seamless Data (ESD) product, spanning the global land surface from 2000 to 2024, is publicly accessible via the iEarth platform at https://data-starcloud.pcl.ac.cn/iearthdata/64. The dataset is organized by year and adheres to the standard MGRS tiling convention to ensure seamless integration with existing Earth Observation workflows. To facilitate the use of these embeddings, all source code and example scripts for downstream task inference are available at https://github.com/shuangchencc/ESD. We encourage the scientific community to utilize these resources for efficient, planetary-scale environmental monitoring and the development of novel geospatial AI applications.

## 7. Conclusion

In this study, we introduced the Embedded Seamless Data (ESD), a first-of-its-kind, ultra-lightweight 30-m Earth embedding database designed for planetary-scale environmental monitoring. By transforming a 25-year record of global land surface reflectance into information-dense, quantized latent vectors, the ESD achieves a transformative ~340-fold reduction in data volume. The entire global land surface for a single year is encapsulated within just 2.4 TB, effectively democratizing petabyte-scale analysis for researchers without access to massive cloud computing infrastructure.

Our evaluation of the ESDNet architecture demonstrates that these embeddings maintain high reconstructive fidelity across the spectral dimension, with an MAE of 0.0130 and consistent stability from 2000 to 2024. Beyond mere compression, the multi-task learning framework ensures that the latent space is semantically organized, significantly outperforming raw reflectance data in downstream tasks such as land-cover classification—achieving an Overall Accuracy of 79.74% compared to 76.92% using raw sensor fusion data—and demonstrating superior performance in few-shot learning scenarios.

The ESD product also exhibits an emergent denoising capability, successfully mitigating residual cloud artifacts and shadows to provide a cleaner, more stable proxy for surface processes. By providing a continuous, longitudinal record that preserves intra-annual phenological dynamics, the ESD offers a robust foundation for tracking decadal shifts in the Earth system. We believe this dataset represents a significant step toward agile Earth system research, enabling global analyses to be executed with the same ease as local studies and fostering new frontiers in geospatial AI.

**Competing interests**. The authors declare that they have no known competing financial interests or personal relationships that could have influenced the work reported in this study.

**Disclaimer**. Publisher's note: Copernicus Publications remains neutral with regard to jurisdictional claims in published maps and institutional affiliations.